\def\year{2022}\relax
\documentclass[letterpaper]{article} % DO NOT CHANGE THIS
\usepackage{aaai22}  % DO NOT CHANGE THIS
\usepackage{times}  % DO NOT CHANGE THIS
\usepackage{helvet}  % DO NOT CHANGE THIS
\usepackage{courier}  % DO NOT CHANGE THIS
\usepackage[hyphens]{url}  % DO NOT CHANGE THIS
\usepackage{graphicx} % DO NOT CHANGE THIS
\usepackage{natbib}  % DO NOT CHANGE THIS AND DO NOT ADD ANY OPTIONS TO IT
\usepackage{caption} % DO NOT CHANGE THIS AND DO NOT ADD ANY OPTIONS TO IT
\DeclareCaptionStyle{ruled}{labelfont=normalfont,labelsep=colon,strut=off} % DO NOT CHANGE THIS
\usepackage{algorithm}
\usepackage{algorithmic}
\usepackage{graphicx}
\usepackage{amsmath}
\usepackage{booktabs}
\usepackage{multirow}
\usepackage{subfigure}
\usepackage{bm}
\usepackage{diagbox}
\usepackage{xcolor}

\usepackage{newfloat}
\usepackage{listings}
\floatstyle{ruled}
\newfloat{listing}{tb}{lst}{}
\floatname{listing}{Listing}
\title{Spatially Focused Attack against Spatiotemporal Graph Neural Networks}

\author{
    Fuqiang Liu, Luis Miranda-Moreno, Lijun Sun\thanks{Corresponding author.}
}

\affiliations{
   McGill University, Montreal, QC, Canada\\
   fuqiang.liu@mail.mcgill.ca, luis.miranda-moreno@mcgill.ca, lijun.sun@mcgill.ca
}

\begin{document}

\maketitle

\begin{abstract}
Spatiotemporal forecasting plays an essential role in various applications in intelligent transportation systems (ITS), such as route planning, navigation, and traffic control and management. Deep Spatiotemporal graph neural networks (GNNs), which capture both spatial and temporal patterns, have achieved great success in traffic forecasting applications. Understanding how GNNs-based forecasting work and the vulnerability and robustness of these models becomes critical to real-world applications. For example, if spatiotemporal GNNs are vulnerable in real-world traffic prediction applications, a hacker can easily manipulate the results and cause serious traffic congestion and even a city-scale breakdown. However, despite that recent studies have demonstrated that deep neural networks (DNNs) are vulnerable to carefully designed perturbations in multiple domains like objection classification and graph representation, current adversarial works cannot be directly applied to spatiotemporal forecasting due to the causal nature and spatiotemporal mechanisms in forecasting models. To fill this gap, in this paper we design Spatially Focused Attack (SFA) to break spatiotemporal GNNs by attacking a single vertex. To achieve this, we first propose the inverse estimation to address the causality issue; then, we apply genetic algorithms with a universal attack method as the evaluation function to locate the weakest vertex; finally,  perturbations are generated by solving an inverse estimation-based optimization problem. We conduct experiments on real-world traffic data and our  results show that perturbations in one vertex designed by SA can be diffused into a large part of the graph.
\end{abstract}

\section{Introduction}
Spatiotemporal traffic forecasting has been a long-standing research topic and a fundamental application in intelligent transportation systems (ITS). For instance, with better prediction of future traffic states, navigation apps can help drivers avoid traffic congestion, and traffic signals can manage traffic flows to increase network capacity. Essentially, traffic forecasting can be modeled as a multivariate time series prediction problem for a network of connected sensors based on the topology of road networks. Given the complex spatial and temporal patterns governed by traffic dynamics and road network structure, recent studies have developed various Graph Neural Networks-based traffic forecasting models and achieved great success~\cite{fang2019gstnet,wu2019graph,attention}.

It has been shown in many recent studies that deep learning frameworks are very vulnerable to carefully designed attacks~\citep[see e.g.,][]{gf1,gf2,gf3,gf4,gf5}. This raises a critical concern about the application of spatiotemporal GNNs-based models for real-world traffic forecasting, in which robustness and reliability are of utmost importance. For example, with a vulnerable forecasting model, a hacker can manipulate the predicted traffic states and feed these manipulated values into the downstream application, thus causing severe problems such as traffic congestion and even city-scale breakdown. Despite having superior accuracy, GNNs-based traffic prediction models are also facing great cyber-security challenges in practice. It remains a critical question to understand and evaluate the vulnerability of these models.

\begin{figure}[!t]
    \centering
    \includegraphics[width=0.75\columnwidth]{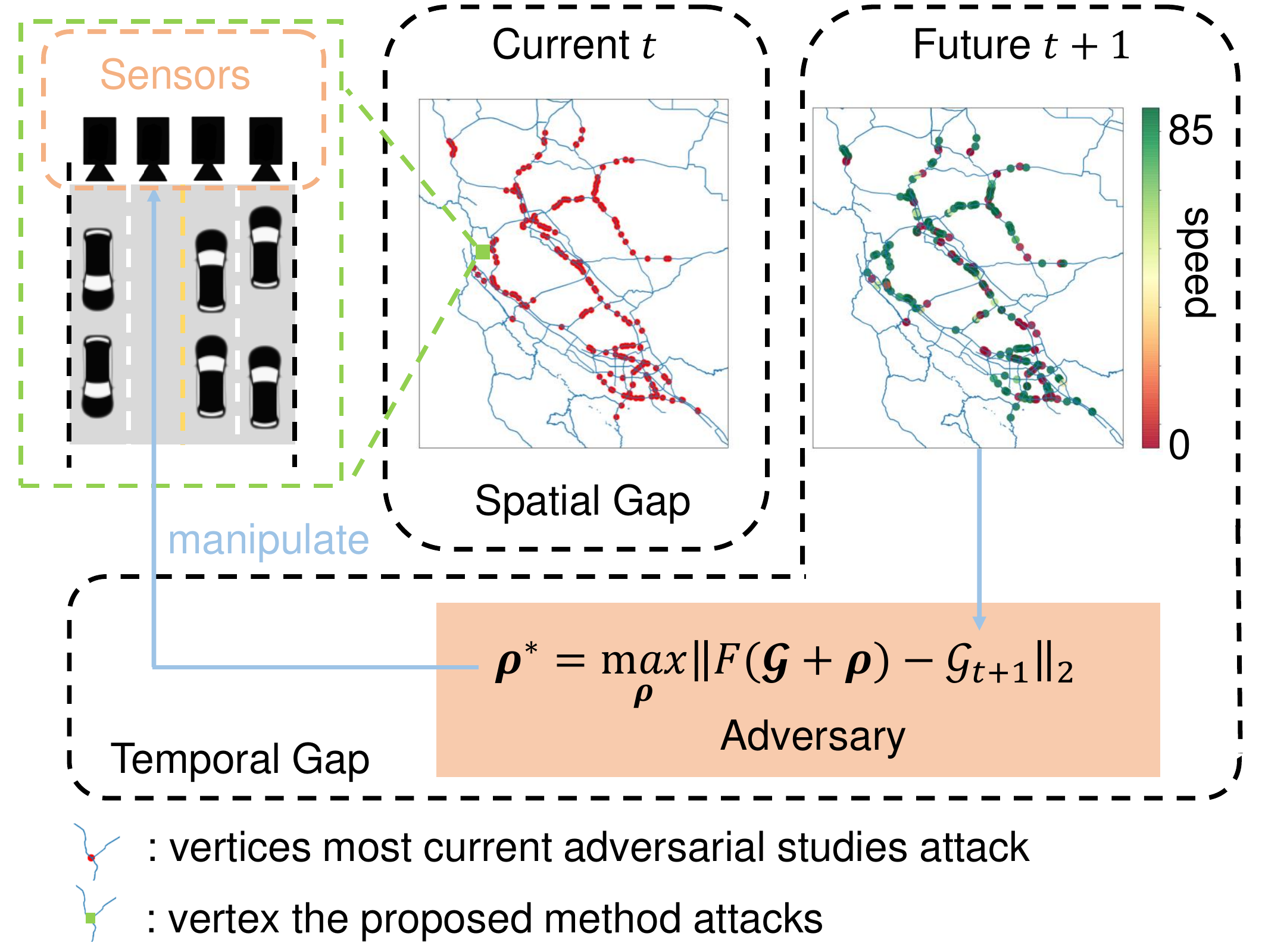}
    \caption{Temporal and spatial gaps when applying current adversarial studies to spatiotemporal forecasting.} %Adversary refers to the agent who generates perturbations to fool the spatiotemporal forecasting models.}
    \label{fig:gaps}
\end{figure}

However, current adversarial works cannot be directly applied to evaluating the vulnerability of GNNs-based spatiotemporal forecasting because of the temporal and spatial gaps shown in Figure~\ref{fig:gaps}. In the following of this paper, we refer to ``sensors'' in a ``road network'' as ``vertices'' of a ``graph'' in GNNs and use two terms interchangeably. First, current adversarial works like adversarial attacks against recurrent neural network (RNN)~\cite{attackrnn1,attackrnn2,attackrnn3} rely on the ground truth to generate perturbations, while the future state---the forecasting models' ground truth---is inaccessible in traffic forecasting applications. For instance, the traffic condition at 11:35 am can not be detected by sensors at 11:30 am. Current adversarial works require the unavailable information at 11:35 am to generate perturbations when they are utilized to fool spatiotemporal forecasting models at 11:30 am. Thus these ground truth-based adversarial models cannot work anymore. We refer to this challenge as the ``temporal gap" (see right and bottom part of Figure~\ref{fig:gaps}). Second, in real-world ITS applications, sensors are deployed on a large-scale road network. Following current adversarial studies, one assumes that all sensors can be manipulated at the same time (see red dots in Figure~\ref{fig:gaps}). However, this assumption is is unrealistic as it is impossible for the hacker vehicle to poison all sensors in such a large-scale road network. We refer to this challenge as the ``spatial gap". More discussions on why current adversarial works cannot be directly applied to attacking GNNs-based forecasting models are detailed in Section~\ref{sec:relatedwork}. Overall, it remains unclear how vulnerable these GNNs-based spatiotemporal forecasting using existing attack frameworks.

The goal of this paper is to understand and examine the vulnerability and robustness of GNNs-based spatiotemporal forecasting models. In doing so, we design a Spatially Focused Attack (SFA) framework to break these forecasting models by manipulating only one vertex in the graph (see the green square in Figure~\ref{fig:gaps}). We first propose Inverse Estimation (IE) to avoid using future ground truth and design and IE-based universal attack mechanism. Then, we utilize the genetic algorithm, of which the evaluation function consists of the proposed universal attack method, to locate the ``weakest” sensor/vertex. Here the weakest vertex refers to the vertex that will it will cause maximum damage to the forecasting models when being attacked. Finally, we generate perturbations by solving an optimization problem. It should be noted that the proposed method does not require future information in designing perturbations. Following the proposed SFA framework, one hacker can break forecasting models by poisoning just one sensor in a large-scale road network. Thus, SFA is a realistic solution to evaluate the robustness and vulnerability of spatiotemporal forecasting models for real-world applications.
%It should be noted that poisoning all vertices in real-world traffic forecasting applications is impossible because of the graph's large-scale. For instance, the graph used in traffic forecasting applications generally covers 1000 square kilometers, and it is unrealistic to organize harker vehicles to poison all vertices in such a large-scale road network. By contrast, one harker vehicle can break forecasting models by poisoning just one vertex based on the proposed OVA. Hence, the proposed one-vertex attack is a realistic solution to evaluate the robustness and vulnerability of spatiotemporal forecasting models deployed in real-world applications.

To prove the effectiveness of the proposed SFA method, we test it on two spatiotemporal traffic datasets with three different Spatiotemporal GNNs, including STGCN \citep{yu2017spatio}, DCRNN \citep{li2017diffusion} and Graph Wavenet \citep{wu2019graph}. Our results show that SFA can cause at least 15\% accuracy drop, and there are about 10\% sensors severely impacted with the boundary of speed variation limited to 15 km/h. The main contributions of this paper can be summarized as follows.
\begin{itemize}
    \item We propose a novel Spatially Focused Attack (SFA) method to find the weakest vertex and break the forecasting model by poisoning single one vertex. To the best of our knowledge, this is the first vulnerability study on GNNs-based spatiotemporal forecasting models by poisoning only one vertex.
    \item We propose to use inverse estimation to avoid using future ground truth when computing perturbations.
    \item We study the effectiveness of the proposed SFA with extensive experiments on real-world datasets.
\end{itemize}
%\begin{itemize}
%    \item To the best of our knowledge, this is the first study on attacking Spatiotemporal GNNs by poisoning only one vertex.
%    \item We proposed a novel OVA method that is able to find the weakest vertex and generate optimal adversarial perturbations.
%    \item We empirically study the effectiveness of the proposed method with multiple experiments on real-world data sets.
%\end{itemize}

%\begin{figure}
%\centering
%\includegraphics[width=8.5cm]{fig.pdf}
%\caption{(a) the spatiotemporal prediction/attack scheme; (b) the perturbation of one vertex attack; (c) the error map of the network-scale prediction under the one vertex attack; (d) the normal traffic condition; (e) possible traffic condition when the prediction-based control system is attacked. In (b, c), the depth of red represents the difference from ground truth; in (d, e), color indicates traffic speed and red corresponds to traffic congestion}
%\label{fig:scheme}
%\end{figure}

\section{Related Work}\label{sec:relatedwork}
\subsubsection{One Pixel Attack for Fooling Deep Neural Networks.}
One pixel attack~\cite{onepixel} utilizes Differential Evolution (DE) to generate the perturbation to poison one pixel in images and then fools CNNs. However, one-pixel attack requires the ground truth to compute perturbations, which means applying one-pixel attack still face the temporal gap. Moreover, images are regular-structured and there exist no temporal variations in one-pixel attack. One-pixel attack' poisoning positions vary in different frames, which is not applicable to spatiotemporal forecasting domains. The above features prevent us from directly using one-pixel attack on spatiotemporal forecasting models.

\subsubsection{Adversarial Attacks against Time Series Analysis.} Some previous works~\cite{attackloadforecasting,attackloadforecasting2,attackarima,attacktimeseries} propose adversarial attack methods against autoregressive models or time series classification models. Essentially, these works only consider univariate time series. Different from these works, we focus on multivariate time series generated from a complex spatial domain/network. The input of spatiotemporal GNNs is a dynamic graph rather than regular matrices or sequences. We take the spatial correlation into consideration, which is overlooked in previous studies.

\subsubsection{Adversarial Attacks against Graph Neural Networks.}
Many studies~\cite{adversarialgraph,adversarialgraph3,zhang2020gnnguard,you2020graph} utilize reinforcement learning (RL), meta learning, or genetic algorithm to fool GNNs in vertex, edge, and graph classification tasks by tuning the graph topology. Still, these studies involve no temporal variations in their graphs, and they mainly focus on the spatial pattern. These models cannot be applied to fool spatiotemporal forecasting models because of the lack of temporal correlation. In particular, attacking spatiotemporal forecasting models deployed in real-world applications by graph topology-based attack methods~\cite{adversarialgraphmeta,adversarialgraph2} are unrealistic, because tuning graph topology corresponds to tuning the structure of sensor networks (i.e., road networks) that collects spatiotemporal data continuously. %Any modification on sensors in real-world applications can be easily sensed by the sensor network manager.

%\textbf{Adversarial Attacks against Recurrent Neural Network.} Recent studies~\cite{attackrnn1,attackrnn2,attackrnn3} demonstrated RNN classifiers were vulnerable to adversarial sequences. These adversarial works require the ground truth to compute adversarial sequences. Because of the forecasting applications' causality, the future ground truth is unavailable. Besides, these works focus on regular vectors or matrices, rather than irregular graphs. Hence these adversarial sequence generation models cannot be directly applied to attack spatiotemporal GNNs-based forecasting models.

\section{Preliminary}
%\subsection{Spatiotemporal Forecasting}

Traffic state data collected from a sensor network is often represented a time varying graph, which encodes both spatial and temporal information. In general, the spatiotemporal sequences can be represented as $\mathcal{G}_{t}=\left\{\mathcal{V}_{t},\mathcal{E},W\right\}$, where $\mathcal{E}$ is the set of edges in the graph, $\mathcal{W}$ is the weighted adjacency matrix in which every element describes the spatial relationship between different sensors, $\mathcal{V}_{t}=\left\{v_{1,t},\ldots,v_{n,t}\right\}$ is the set of state values (e.g. traffic speed or traffic volume) collected from sensors on timestamp $t$, and $n$ is the number of sensors~\cite{gconv}.  For multi-step spatiotemporal forecasting, future states are estimated as
\begin{equation}~\label{equ:forecast}
\left\{\mathcal{G}^{*}_{t+M},...,\mathcal{G}^{*}_{t+1}\right\} = F\left(\left\{\mathcal{G}_{t},...,\mathcal{G}_{t-N+1}\right\}\right),
\end{equation}
where $\mathcal{G}^{*}_{t}$ denotes the prediction of the states at time $t$. Previous states from $t-N+1$ to $t$ are fed into a forecasting model $F$ that outputs predictions of future states from $t+1$ to $t+M$. In general, we have $M \le N$. The above process is customarily called sequence-to-sequence (seq2seq) forecasting. Most spatiotemporal forecasting models output a single future state, which will be in turn fed as input into the model to forecast the next state. This process is named as the recursive multistep forecasting, and future states are computed as
\begin{equation}~\label{equ:recursive}
\left\{
\begin{aligned}
\mathcal{G}^{*}_{t+1} &= F\left(\left\{\mathcal{G}_{t},\mathcal{G}_{t-1},...,\mathcal{G}_{t-N+1}\right\}\right)\\
\mathcal{G}^{*}_{t+2} &= F\left(\left\{\mathcal{G}^{*}_{t+1},\mathcal{G}_{t},...,\mathcal{G}_{t-N+2}\right\}\right)\\
\vdots  \\
\mathcal{G}^{*}_{t+M} &= F\left(\left\{\mathcal{G}^{*}_{t+M-1},\mathcal{G}^{*}_{t+M-2},...,\mathcal{G}_{t-N+M}\right\}\right).\\
\end{aligned}
\right.
\end{equation}

State-of-the-art forecasting models, $F$, are generally constructed based on spatiotemporal GNNs~\citep[see e.g.,][]{li2017diffusion,wu2019graph,yu2017spatio,attention}, consisting of both spatial layers and temporal layers. In general, these models use gated linear unit (GLU) or Gated-CNN~\cite{gruuu,gatedcnn} as the temporal layer to capture the temporal patterns embedded in the spatiotemporal sequence, and the Graph-CNN~\cite{gconv,graphcnn} is used as spatial layers to capture the spatial patterns. In this paper, we concentrate our adversarial studies on recursive multistep spatiotemporal forecasting models. However, the analysis can be easily extended to seq2seq-based multistep forecasting.

\section{Methodology}

In this section, we detail the proposed SFA framework, which essentially consists of three components. We first propose to use Inverse Estimation (IE) as a potential solution to address the temporal gap. Second, we derive a solution to locate the weakest vertex to avoid manipulating all sensors. Finally, an optimization model is introduced to SFA to compute and design the perturbations used to fool spatiotemporal forecasting models.

\subsection{Inverse Estimation}~\label{sec:ie}
%\subsubsection{Adversarial Attack}
In the domain of computer vision, adversarial attacks aim at fooling a machine learning-based classifier to misclassify objects with undetectable modifications. When it comes to spatiotemporal forecasting, adversarial attacks is to add unnoticeable perturbations into historical time series such that the forecasting models begin to generate bad predictions that are far away from the ground truth. We formulate this goal as the the following optimization problem:
\begin{equation}~\label{equ:adv}
\begin{split}
\mathop{\max}_{\bm{\rho}} & \left\|F\left(\bm{\mathcal{G}}+\bm{\rho}\right)-{\mathcal{G}}_{t+1}\right\|_{2} \\
{\rm s.t.}\   &\rho_{i}^2 \le \xi,
\end{split}
\end{equation}
where $\bm{\mathcal{G}}=\left\{\mathcal{G}_{t},...,\mathcal{G}_{t-N+1}\right\}$ denotes the input graph sequence, $\mathcal{G}_{t+1}$ denotes the future traffic state that is the ground truth of the forecasting model, $\bm{\rho}=\{\rho_{t},...,\rho_{t-N+1}\}$ denotes a collection of perturbations, and $\xi$ denotes the pre-specified constant to constrain the perturbation scale in order to make the perturbations unnoticeable.

%Adversarial attack aims at fooling a DNN model by an unnoticeable perturbation, which can be generally formed as
%\begin{equation}~\label{equ:adv}
%\begin{split}
%\mathop{\min}_{\bm{\rho}}\ \left\| \bm{\rho}\right\|_{2} \\
%{\rm s.t.}\ F\left(X+\rho\right) \neq L,
%\end{split}
%\end{equation}
%where $X$ denotes the input and $\rho$ denotes the perturbation.

%where $\bm{\mathcal{G}}$ denotes the input graph sequence and $\bm{\mathcal{G}}=\{\mathcal{G}_{t},...,\mathcal{G}_{t-N+1}\}$, $\bm{\rho}$ denotes perturbations and $\bm{\rho}=\{\rho_{t},...,\rho_{t-N+1}\}$, $\rho_i$ denotes the perturbation on timestamp $i$, $\| \cdot\|_{p}$ denotes $\ell_p$-norm, and $\xi$ denotes the pre-defined constant to constrain the perturbation scale. In real-world traffic applications, $\xi$ control the hacker's driving behavior to balance the attack performance and detection avoidance. The goal of Eq.~\ref{equ:newadv} is to mislead spatiotemporal GNNs to generate false forecasting by adding perturbations.

However, it is difficult to directly solve the optimization model in ~\eqref{equ:adv}. As an alternative, we design a proxy optimization problem by integrating the constraint into the objective function:
\begin{equation}~\label{equ:newadv}
\mathop{\max}_{\bm{\rho}}  \left\|F\left(\bm{\mathcal{G}}+\bm{\rho}\right) - {\mathcal{G}}_{t+1} \right\|_2 - \alpha  \sum_{i=t-N+1}^{t} \max\left(0,\rho_i^2-\xi\right),
\end{equation}
where $\alpha$ denotes the penalty factor.
%where $\bm{\mathcal{G}}=\left\{\mathcal{G}_{t},...,\mathcal{G}_{t-N+1}\right\}$ denotes the input graph sequence, $\bm{\rho}=\{\rho_{t},...,\rho_{t-N+1}\}$ denotes perturbations, $\rho_i$ denotes the perturbation on timestamp $i$, $\alpha$ denotes the penalty factor, and $\xi$ denotes the pre-defined constant to constrain the perturbation scale.
Even though the regularization term in Eq.~\eqref{equ:newadv} is a soft constraint compared to the original constraint $\left\| \rho\right\|_{2} \le \xi$, it can still strictly force the perturbation's scale less than the upper bound, $\xi$, by setting the penalty factor to be a large value to make the scale penalty term much larger than the first term in Eq.~\eqref{equ:newadv}. In real-world traffic applications, $\xi$ is used to balance the attack performance and degree of observability. %The goal of Eq.~\ref{equ:newadv} is to mislead spatiotemporal GNNs to generate false forecasting by adding unnoticeable perturbations.

However, the fact is that the ground truth of a forecasting model, $\mathcal{G}_{t+1}$ in Eq.~\eqref{equ:newadv}, is unavailable at time $t$. This issue is referred to as the ``temporal gap" in Figure~\ref{fig:gaps}. Due to the inevitable usage of the future information, fooling spatiotemporal GNNs as Eq.~\eqref{equ:newadv} with the simple optimization model becomes unrealistic. A simple solution to address this issue is to directly use the most recent observations as the forecasting. This is equivalent to approximating the complex GNNs with a ``most recent'' forecaster:
\begin{equation}~\label{equ:de}
\mathcal{G}_{t+1} \leftarrow \mathcal{G}_{t}.
\end{equation}
However, perturbations computed by the direct estimation may not be effective enough to cause significant performance drop. The reason is the direct estimation involves great errors in the perturbation computation, which is discussed in Section~\ref{sec:ie_de}. In addition, the maximization optimization problem is still difficult to solve given the large search space.

%\subsection{Universal adversarial attack against spatiotemporal GNNs}
%In this section, we point out the form of adversarial attack against the spatiotemporal forecasting, and outline the gap between attacking spatiotemporal GNNs and attacking CNNs or normal GNNs. Then we propose the inverse estimation to fill the gap. Finally, we design the universal adversarial attack against spatiotemporal GNNs. The proposed universal adversarial attack will work as the evaluation function to locate the weakest vertex in Section~\ref{subsection:weakest}.

%\subsubsection{Adversarial attacks against spatiotemporal forecasting}
%Because of spatiotemporal sequence's causality, we cannot access the future condition that works as forecasting models' ground truth.

%\subsubsection{Inverse estimation}\label{sec:inverse}

To address the said issue, we propose an Inverse Estimation scheme to offer a simple but clear direction to the optimization model. We introduce the concept of ``opposite state'' and transform the maximization problem in \eqref{equ:newadv} into a minimization problem, whose goal is to fool spatiotemporal GNNs to generate predictions opposite to the ground truth:
\begin{equation}~\label{equ:newnewadv}
\mathop{\min}_{\bm{\rho}}  \left \|F \left(\bm{\mathcal{G}}+\bm{\rho} \right) - \tilde{\mathcal{G}}_{t+1} \right\|_2 + \alpha  \sum_{i=t-N+1}^{t} \max \left(0,\rho_i^2-\xi\right),
\end{equation}
where $\tilde{\mathcal{G}}_{t+1}$ denotes the ``opposite state'' of $\mathcal{G}_{t+1}$. Perturbations can be generated more effectively by solving Eq.~\eqref{equ:newnewadv}. The above idea is similar to targeted attacks~\cite{survey}. However, classical targeted attacks still utilize the ground truth in perturbation computations.

%The constraint in Eq.~\ref{equ:newadv} is replaced with a regularization term in Eq.~\ref{equ:newnewadv} to constrain the perturbation scale. The penalty factor $\alpha$ is set as 100 to make sure the scale penalty term is much larger than the first term in Eq.~\ref{equ:newnewadv} so that the scale of the computed perturbation is strictly forced.

To avoid using future information, the opposite of future state, $\tilde{\mathcal{G}}_{t+1}$, is estimated by computing the opposite of the most recent state:
\begin{equation}~\label{equ:ie}
\tilde{\mathcal{G}}_{t+1} \leftarrow \tilde{\mathcal{G}}_{t}=\left\{\tilde{\mathcal{V}}_{t},\mathcal{E},W\right\},
\end{equation}
where $\tilde{\mathcal{V}}_{t}=\left\{\tilde{v}_{1,t},...,\tilde{v}_{n,t}\right\}$ denotes a collection of state values opposite to these collected from sensors. Taking the traffic speed as an example, when the condition is ``congested/low speed", its opposite state should be ``free/high speed". In this case, we compute $\tilde{v}_{i,t}$---the the opposite of $v_{i,t}$---using two distinct values:
\begin{equation}~\label{equ:computeopposite}
\tilde{v}_{i,t}=\left\{
		\begin{aligned}
		\max\left(\mathcal{V}\right), & \quad v_{i,t} < {\rm mid}\left(\mathcal{V}\right), \\
		\min\left(\mathcal{V}\right), &\quad v_{i,t} \ge {\rm mid}\left(\mathcal{V}\right),
	\end{aligned}
	\right.
\end{equation}
where ${\rm mid}\left(\mathcal{V}\right)$, $\max\left(\mathcal{V}\right)$, and $\min\left(\mathcal{V}\right)$ represent the mean, maximum, and minimum value of the spatiotemporal dataset, respectively. We examine the effectiveness of this approach in the experiment section with real-world datasets.

%We proposed the inverse estimation to avoid using the ground truth in perturbation computation. However, perturbations generated by solving Eq.~\eqref{equ:newnewadv} should be added into the entire traffic graph, which is unrealistic because of the large size of real-world road network. %The spatial gap is not filled.

\subsection{Locating the Weakest Vertex}\label{sec:weakest}

In this section, we introduce a solution to avoid manipulating (introducing perturbations on) all sensors in the road network. The key idea is to identify the most vulnerable sensor, which produces the largest accuracy drop to the whole sensor network when being attacked. To achieve this goal, we first solve a universal attack problem by making the perturbation static. Thus, the universal attack implies that the perturbation is consistent and independent from the input. The universal perturbation is computed by solving the following optimization problem
\begin{equation}~\label{equ:newnewnewadv}
\mathop{\min}_{\rho_{u}} \left\|F\left(\bm{\mathcal{G}}+\{\rho_u\}\right) - \tilde{\mathcal{G}}_{t+1} \right\|_2 + \alpha \cdot  \max \left(0,\rho_u^2-\xi\right),\\
\end{equation}
where $\rho_{u}$ denotes the universal perturbation. After the universal perturbation is generated, there is no need to update it when new data come. It should be noted that this universal perturbation can only break forecasting models by poisoning all sensors. Thus it is unrealistic to apply it to real-world forecasting applications. Nevertheless, this universal perturbation will be used to define and quantify sensor weakness.

% the IE-based universal attack against spatiotemporal forecasting is proposed; second we mathematically define the ``weakness" of a vertex based on the proposed universal attack; finally how to locate the ``weakest" vertex is detailed.

We define the ``weakness" score of the $j$th sensor as the number of influenced/affected sensors when sensor $j$th is attacked by the proposed universal perturbation. Specifically, the weakness is computed as
\begin{equation}\label{equ:weak}
weak_{j,t}= \left \|K_{\theta} \left\{ F\left(\bm{\mathcal{G}}+M_{j}\cdot \rho_{u}\right)- \mathcal{G}_{t+1} \right\}\right\|_{0}, \\
\end{equation}
where $M_j\cdot \rho_{u}$ denotes that all elements except the one corresponding to the $j$th sensor, and $K_{\theta}\left\{{\cdot}\right\}$ denotes an element-wise filter to set elements whose absolute value is smaller than $\theta$ to 0. As Eq.~\eqref{equ:weak} shows, the weakness is actually time-dependent. By collecting the weakness in a time frame, a weakness vector can be formed. The $l$2-norm of the vector can be regarded as the time-invariant weakness for a vertex. Thus, for a sensor $j$, a greater weakness value suggests that more sensors will be influenced if it is attacked. We will attack the vertex with the largest ``weakness" value. For traffic forecasting applications, we consider the vertex where the prediction error is greater than 5 km/h as the influenced vertex, and $\theta$ is set to 5 consequently.

\begin{figure}[!t]
    \centering
    \includegraphics[width=0.75\columnwidth]{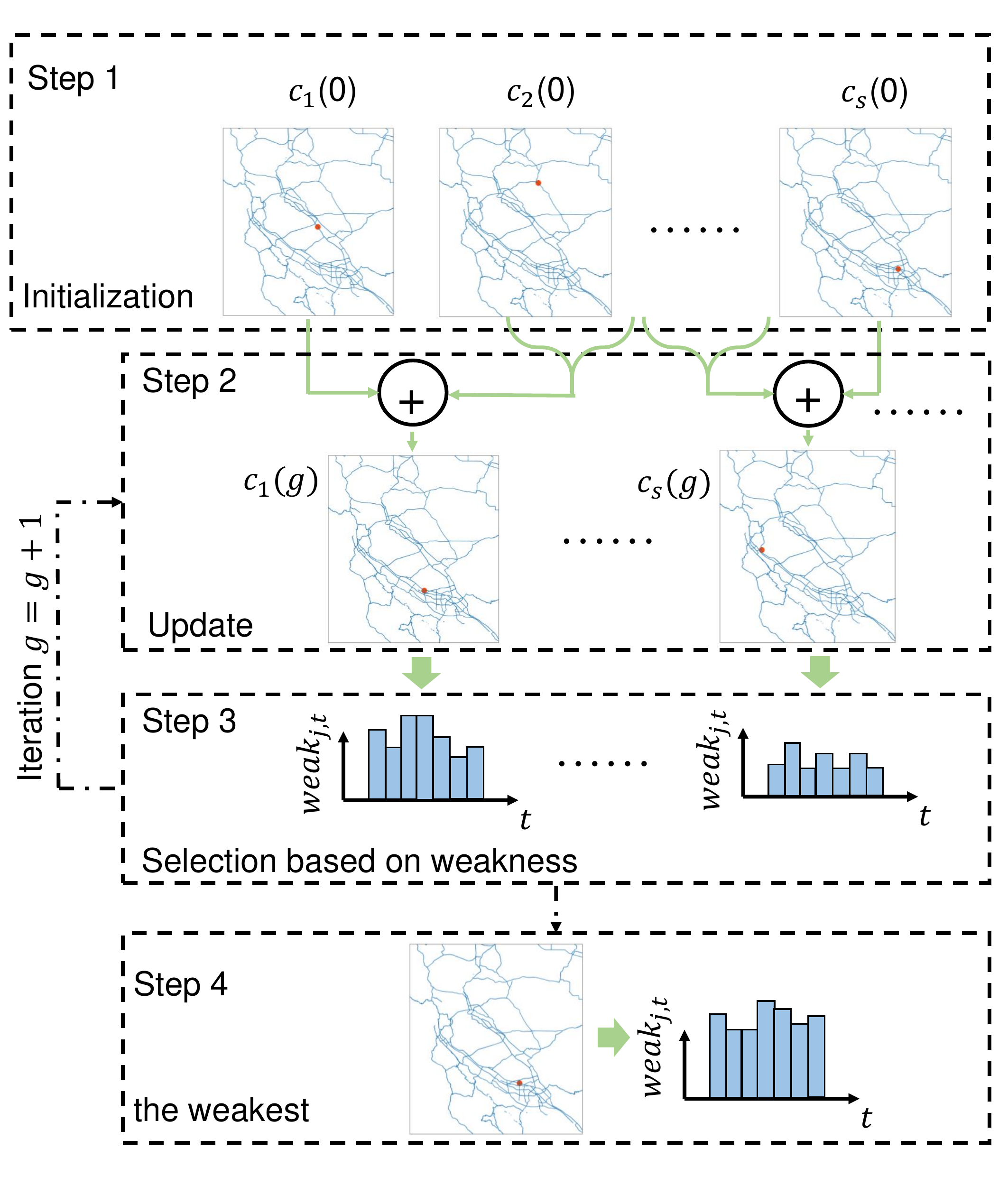}
    \caption{Locating the weakest vertex. Please note the red dot refers to vertex to poison.}
    \label{fig:locate}
\end{figure}

A possible method to locate the weakest vertex is the complete traversal algorithm. However, this method is time consuming. To reduce the time cost, we utilize the genetic algorithm to locate the weakest vertex, which is shown as follows. The genetic process is presented schematically in Figure~\ref{fig:locate}.
\begin{itemize}
    \item First, the initial candidate set is generated with $s$ sensors with the most edges.
    \item Second, the updated set consists of $s$ new candidates and they are computed as
    \begin{equation}\label{equ:DE}
        c_i(g+1)=c_{r1}(g)+p(c_{r2}(g)-c_{r3}(g)),
    \end{equation}
    where $c_{i}$ denotes the position (longitude and latitude) of the $i$th vertex, $g$ denotes the $g$th iteration, $r1$, $r2$, and $r3$ are random numbers with different values, and $p$ is set to  0.5 empirically.
    \item Third, compare the weakness of updated candidates with the previous candidate set, then keep only $s$ candidates with the largest weakness value.
    \item Fourth, repeat the second and third step until the candidate set is consistent or $g$ is sufficient. Select the weakest vertex to attack. It should be noted that the bound of $g$ controls the trade-off of the effectiveness and efficiency of the solution. The larger bound represents the proposed solution is much closer to the complete traversal algorithm.
\end{itemize}

%\begin{algorithm}[tb]
%\caption{Locating the weakest vertex}
%\label{alg:algorithm}
%\textbf{Input}: Your algorithm's input\\
%\textbf{Parameter}: Optional list of parameters\\
%\textbf{Output}: Your algorithm's output
%\begin{algorithmic}[1] %[1] enables line numbers
%\STATE Let $t=0$.
%\WHILE{condition}
%\STATE Do some action.
%\IF {conditional}
%\STATE Perform task A.
%\ELSE
%\STATE Perform task B.
%\ENDIF
%\ENDWHILE
%\STATE \textbf{return} solution
%\end{algorithmic}
%\end{algorithm}

\subsection{Spatially Focused Attack}
Once the weakest vertex is located, spatially focused attack is proposed to fool the spatiotemporal forecasting model by poisoning only the selected vertex. The perturbation is computed by solving the following optimization problem:
\begin{equation}~\label{equ:onevertex}
\begin{split}
\mathop{\min}_{M_{J}\cdot\bm{\rho}} \left\|F\left(\bm{\mathcal{G}}+M_{J}\cdot\bm{\rho}\right) - \tilde{\mathcal{G}}_{t}\right\|_2+\alpha\cdot \mathcal{R},
\end{split}
\end{equation}
where $M_J\cdot\bm{\rho}$ is the generated one vertex perturbation, $J$ denotes the index of the weakest vertex, and $\mathcal{R}= \max\left(0,(M_J\cdot\bm{\rho})^2-\xi\right)$ denotes the regularization term to control the scale of the generated one vertex perturbation.
%It should be noted that $\|M_J\cdot \rho_i\|_{0}\leq 1$.% is a single value rather than a matrix or a vector, and $\sqrt{\xi}$ denotes the boundary of speed variation in real-world traffic forecasting applications.

%Poisoning the weakest vertex in a graph with the carefully designed perturbation can fool the spatiotemporal GNNs-based traffic forecasting system.

%\begin{equation}~\label{equ:onevertex}
%\begin{split}
%\mathop{\arg\min}_{M_{J}\cdot\{\rho_{t},...,\rho_{t-N+1}\}}\ \ \| \mathbb{G}_{t+1} - %\tilde{\mathcal{G}}_{t}\|_2+\alpha\cdot \mathcal{R}_{onevertex}\\
%\mathbb{G}_{t+1}=F(\{\mathcal{G}_{t},...,\mathcal{G}_{t-N+1}\}+M_{J}\cdot\{\rho_{t},...,\rho_{t-N+1}\})\\
%\mathcal{R}_{onevertex}=\sum_{i=t-N+1}^{t} max(0,(M_J\cdot\rho_i)^2-\xi)\\
%\end{split}
%\end{equation}

Different from adversarial attack methods as in Eq.~\eqref{equ:newnewadv} and Eq.~\eqref{equ:newnewnewadv} that manipulates all sensors, the proposed SFA poisons only one sensor in the road network in order to achieve the largest accuracy drop. Therefore, SFA provides a more realistic and reasonable framework for implementing real-world attacks. In reality, perturbation on a selected sensor can be introduced by making a hacker vehicle drive by or by controling the network communication to generate fake sensor readings. Overall, we consider SFA an essential and meaningful attack strategy and thus it can be used to evaluate the robustness and vulnerability of different GNNs-based traffic forecasting models.

% The one vertex perturbation denotes a vehicle's speed shift. If a hacker vehicle's speed varies following the perturbation computed from Eq.~\eqref{equ:onevertex}, it will fool the entire traffic forecasting system, not only at the vertex where the hacker is, but also at other sensors and even sensors far away from the attacked vertex.

%One of experiments in Section~\ref{sec:evaluation} shows one vertex attack in a real world road network, and its details are shown as Fig.~\ref{fig:attackmap}~\ref{fig:input}~\ref{fig:55}~\ref{fig:205}.

%\subsection{Effectiveness discussion}~\label{sec:insight}
% Temporal and spatial gaps prevent ITS managers from applying current well-developed adversarial studies to evaluating spatiotemporal forecasting models.

\section{Evaluation and Results}\label{sec:evaluation}

In this section, we evaluate the proposed SFA framework on two traffic datasets, namely \textbf{PeMS} and \textbf{METR-LA(S)}. PeMS consists of traffic speed data from 200 detectors of Caltrans Performance Measurement System (PeMS) and METR-LA(S) also registers traffic speed data for 100 detectors on the highways of Los Angeles County. These two datasets have been widely used as a benchmark to assess spatiotemporal GNN models. Our experiments are conducted on an NVIDIA Tesla V100 GPU.

We test three spatiotemporal GNNs-based forecasting models including STGCN~\cite{yu2017spatio}, DCRNN~\cite{li2017diffusion}, and Graph WaveNet~\cite{wu2019graph}. Each dataset is split into 3 subsets: 70\% for training, 10\% for validation, and 20\% for test. All parameters are as the same as in the original studies except that we set the number of input and output channels to be consistent with the number of detectors/sensors. We use the validation set to locate the weakest sensor, and generate SFA perturbations in real-time for the test set. As for evaluation metrics, we introduce three metrics to quantify the effectiveness of attacks:
\begin{itemize}
    \item \textbf{MAPE Increase (MAPEI)}: MAPE is a measure of prediction accuracy and smaller MAPE represents better predictions. An increase in MAPE thus translates into a decrease in the prediction accuracy.
    \item \textbf{Normalized MAPE Increase (NMAPEI)}: The ratio between MAPEI and MAPE before the attacks.
    \item \textbf{$k\%$-Impacted Vertices ($k$\%-IV)}: The number of vertices with NMAPEI greater than $k$\%.
\end{itemize}

\subsection{Effectiveness of Inverse Estimation}\label{sec:ie_de}
We compare the effectiveness of perturbations generated from the inverse estimation (Eqs.~\eqref{equ:newnewadv} and ~\eqref{equ:ie}) with the direct estimation (Eqs.~\eqref{equ:newadv} and ~\eqref{equ:de}). It should be noted that in this experiment we apply perturbations on all sensors instead of attacking only one sensor. We conduct the analysis on 15 min-head traffic prediction on the PeMS data with STGCN as the base model.

\begin{figure}[!t]
    \centering
    \includegraphics[scale=0.3]{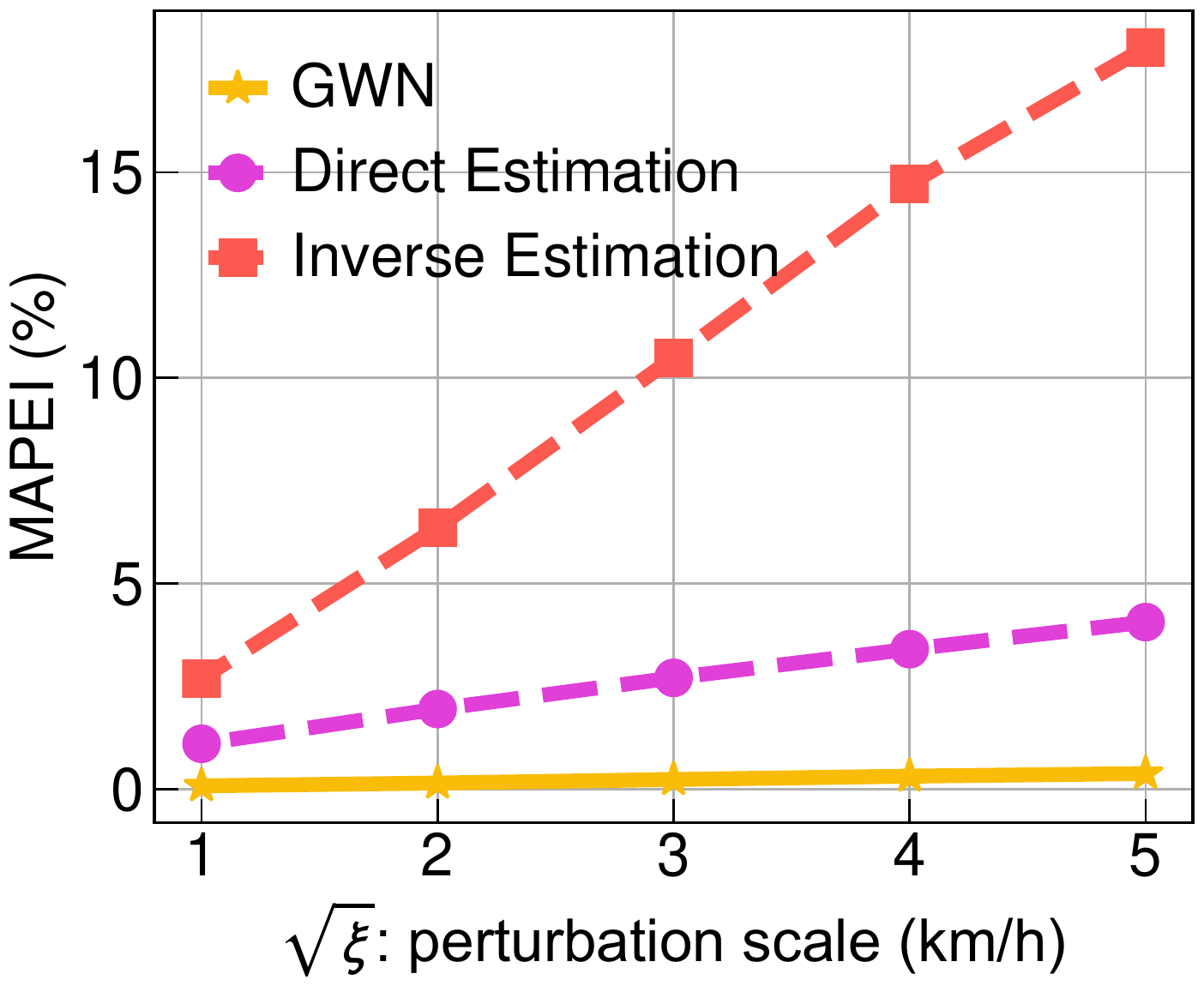}
    \caption{Comparison between the inverse estimation and the direct estimation.}
    \label{fig:ie_de}
\end{figure}

Figure~\ref{fig:ie_de} shows the performance comparison between the proposed inverse estimation and the baseline, direct estimation. As can be seen, perturbations generated from inverse estimation can cause the forecasting forecasting model's accuracy drop greater compared with perturbations generated from direct estimation. Inverse Estimation outperforms directly estimating the future ground truth because the proposed IE can feed less errors into the perturbation computation process. Both strategies cannot achieve perfect estimation and their estimation errors can impact the effectiveness of perturbations. The proposed inverse estimation is a binary estimation and it is much easier than estimating a continuous value. Thus the proposed IE involves less errors in perturbation computations, which in turn, as a result, also lead to more effective adversarial examples. Take PeMS for instance, errors fed by IE is small (MAPE: 0.56\%), while errors fed by the direct estimation is large (MAPE: 3.3\%).

%However, IE's estimation errors fed into the perturbation computation is much smaller than errors fed by the direct estimation. Take PeMS for instance, errors fed by IE is small (MAPE: 0.56\%), while errors fed by the direct estimation is large (MAPE: 3.3\%).

%Moreover, for PeMS, we find that estimating the opposite of the ground truth only generates a few errors (MAPE is 0.56\%), while directly estimating the ground truth generates larger errors (MAPE is 3.3\%). Less errors in perturbation computations lead to more effective adversarial examples.

\subsection{Experiments on Hyperparameters}\label{sec:hyper}

We next examine the effect of hyperparameters on locating the weakest sensor. The setting is as same as that in Section~\ref{sec:ie_de}. We set the number of candidates $s$ to 5 and 10, respectively, and record the number of the iterations. We use NMAPEI and computation time to measure and compare the performance of different hyperparameter configurations.

\begin{figure}[!b]
    \centering
    \includegraphics[scale=0.28]{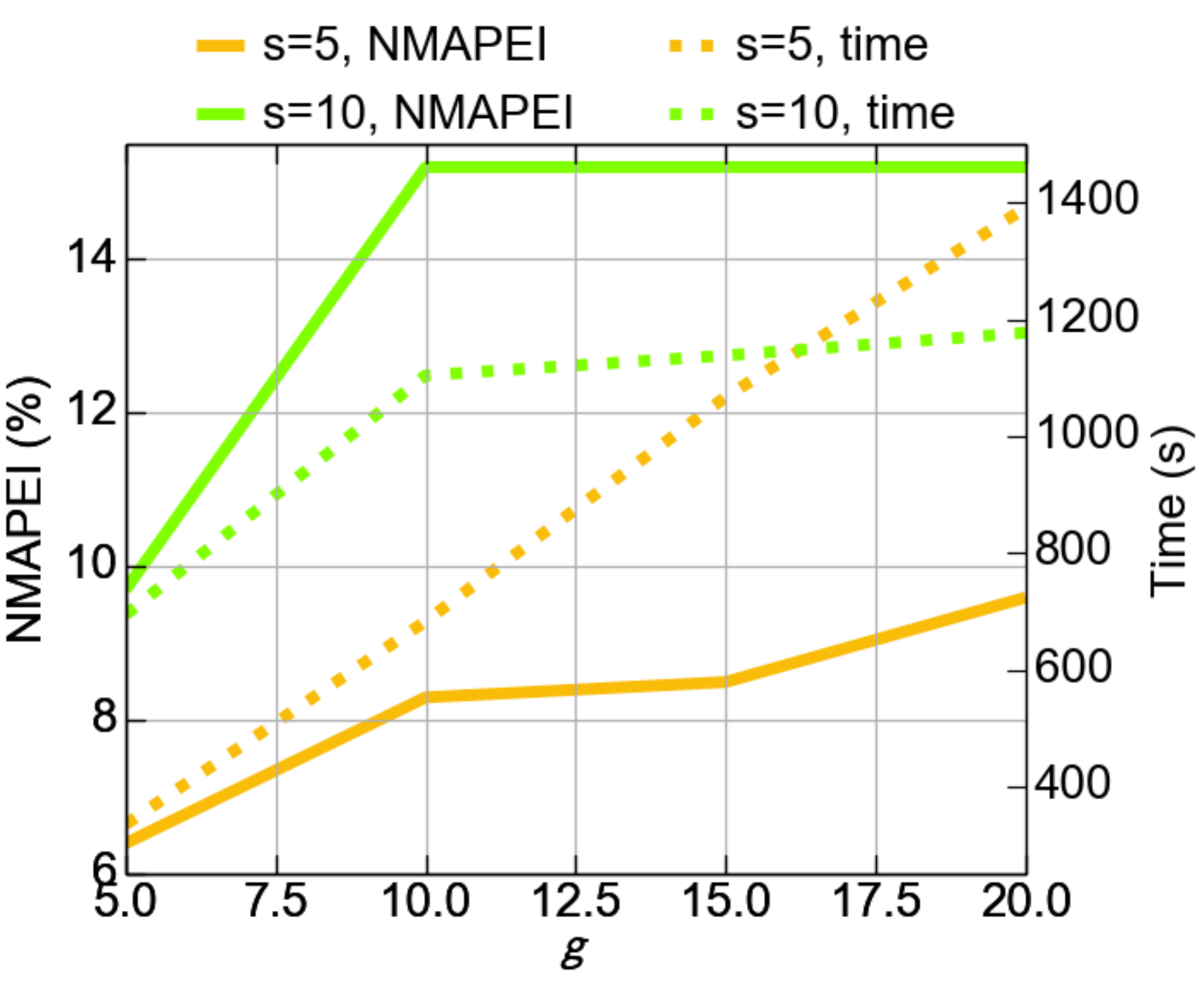}
    \caption{Experiments on hyperparameters of locating the weakest vertex. The number of candidates, $s$, and the iteration, $g$, are studied.}
    \label{fig:hyper}
\end{figure}

From Figure~\ref{fig:hyper}, the computation time generally grows with the increase of the iteration $g$. When $s$ is set to 5, locating the weakest vertex is much harder compared with the case of setting $s$ to 10. A possible reason is that, with a few initial candidates, the proposed strategy tends to converge to a local optimum. Note that when we set $s$ to 10, the proposed strategy breaks the iteration loop. For the following experiments, we set both $s$ and $g$ to 10.

%Section~\ref{sec:hyper} discuss the impact of the hyperparameters including the number of initial candidates, $s$, and the iteration, $g$. Based on the said empirical studies, the number of candidates, $s$, is set as 10 and the bound of the iteration is 10. Besides, Section~\ref{sec:locating weakest} studies the effectiveness and efficiency of the proposed solution on locating the weakest vertex. The proposed weakest vertex locating method is compared with other possible methods. Its performance is close to the ideal complete traversal algorithm and it will save almost 40\% computational cost.

\subsection{Experiments on Effectiveness and Efficiency of Locating the Weakest Vertex}\label{sec:locating weakest}

In this section, we design experiments to demonstrate the effectiveness and efficiency of different strategies in locating the weakest sensor. The setting in Section~\ref{sec:ie_de} is applied to this experiment. We compare the proposed approach with three simple baselines, including (1) locating the vertex with the highest degree (DEG), i.e., the number of connected sensors, (2) locating the vertex with the highest weighted degree centrality (CEN), i.e., row-sum of the weighted adjacency matrix, and (3) locating the weakest vertex by the complete traversal algorithm (CT). After locating the weakest vertex by different strategies, perturbations are computed based on Eq.~\eqref{equ:onevertex} and then fed into STGCN. We evaluate different approaches using NMAPEI, 30\%-IV, and computation time.

%Three locating strategies, namely locating the vertex with the most edges (DEG), locating the vertex with the highest centrality (CEN), and locating the weakest vertex by the complete traversal algorithm (CT), work as baselines.

\begin{table}[!t]
\small
\centering
\begin{tabular}{lrrr}
\toprule
& NMAPEI (\%) & 30\%-IV & time (s) \\
\midrule
DEG & 4.5 &   3 & -\\
CEN & 3.2 &   1 & -\\
CT & \textbf{15.2} &   \textbf{17} & 1795\\
\midrule
Proposed ($s$=10, $g$=10) & \textbf{15.2} &   \textbf{17} & 1104\\
\bottomrule
\end{tabular}
\caption{The effectiveness and efficiency analysis on the proposed strategy to locate the weakest vertex.}\label{table:effectiveness weak}
\end{table}

Table~\ref{table:effectiveness weak} shows the comparison results. As we can see, the proposed solution achieves the same optimal as CT---it identifies the same weakest sensor as the full enumeration. On the other hand, simply poisoning the vertices with highest degree and the highest centrality cannot ensure an effective and efficient attack. A possible reason is that the robustness of sensors is improved by their neighbors because of the local/spatial aggregation mechanism in GNNs. Besides, the proposed strategy can reduce 40\% computation cost compared with the CT.

\subsection{Tradeoff between Attack Performance and the Attack Observability}

%In real-world traffic applications, the generated perturbations represent the hacking vehicle's speed shifts. The parameter $\xi$, which is used to limit the driving behavior, in Eq.~\ref{equ:onevertex} balances the attack performance and detection avoidance.
\begin{table*}[!t]
\small
\centering
\begin{tabular}[]{ l r r r r r r r r r r r r}
\toprule
 &\multicolumn{4}{c}{STGCN} &\multicolumn{4}{c}{DCRNN } &\multicolumn{4}{c}{Grave Wavenet }\\
\cmidrule{2-13}
{$\sqrt{\xi}$ (km/h)} & 5 & 10 & 15 & 20 & 5 & 10 & 15 & 20 & 5&10 & 15 & 20\\
\midrule
5\%-IV & 43 & 71 & 69 &  90& 51 & 75 & 82& 88& 42& 70 & 82 & 91\\
10\%-IV & 14& 52 & 61 & 82& 17 & 59 &65 & 71& 16& 32 & 52 & 73\\
20\%-IV & 4 & 22 & 40 & 46 & 8 & 27 & 45& 50& 3& 25& 41 & 55\\
30\%-IV & 0 & 1 & 17 & 39 & 1 & 1 & 26& 40& 1& 4& 19 & 38\\
40\%-IV & 0 & 1 & 1 & 9 &  0&  0& 0&13 &0&0 & 1 & 16\\
\bottomrule
\end{tabular}
\caption{The relationship between $\sqrt{\xi}$ and the $k\%$-IV.}\label{table:snr-effectiveness}
\end{table*}

In this section, we evaluate examine the effect of perturbation scale on the attack performance. We can consider perturbation scale an indicator for attack observability. A larger perturbation is more likely to be noticed by the user of GNNs. We first propose an experiment to assess how the parameter $\xi$ in Eq.~\eqref{equ:onevertex} influences the effectiveness of the proposed one vertex attack method. In this subsection, 15 min traffic speed forecasting is undertaken by STGCN, DCRNN, and Graph Wavenet that work as the targeted models and the experiment is conducted on META-LA(S). These models are attacked by the proposed SFA with different $\xi$. Note that only one sensor/vertex is poisoned in this experiment.

%\begin{equation}\label{equ:snr}
%SNR=\frac{\sum \|M_J\cdot\mathcal{G}_{t}\|_2}{\sum \|M_J\cdot\rho_{t}\|_2}
%\end{equation}

%\begin{figure}[htbp]
%\begin{subfigure}{.5\textwidth}
%\centering
%\includegraphics[width=4cm]{snrmaped.pdf}
%\caption{MAPEI}\label{fig:maped}
%\end{subfigure}
%\begin{subfigure}{.5\textwidth}
%\centering
%\includegraphics[width=4cm]{snrnmaped.pdf}
%\caption{NMAPEI}\label{fig:nmaped}
%\end{subfigure}
%\caption{MAPEI and NMAPEI with different $\sqrt{\xi}$}
%\end{figure}

%Fig.~\ref{fig:maped} and Fig.~\ref{fig:nmaped} present MAPEI and NMAPEI when these three spatiotemporal GNNs-based forecasting methods are attacked by the proposed one vertex attack method with different $\xi$.
Table~\ref{table:snr-effectiveness} shows the number of impacted vertices (IV) with different $\xi$. When setting $\sqrt{\xi}$ to 20 km/h, we find that around 10\% sensors will have an NMAPEI greater than 40\% and $\sim$90\% sensors will show at least 5\% increase in NMAPEI. This suggests that, when we have a large $\sqrt{\xi}$, the whole network could be severely impacted even attacking only one sensor.  With a small $\sqrt{\xi}$, there are about 50\% sensors are influenced for at least 5\% NMAPE. Based on Table~\ref{table:snr-effectiveness}, we can conclude that perturbations will be effectively diffused from one vertex to most of the graph when we apply GNNs-based spatiotemporal  forecasting models. The greater the perturbation is, the larger the number of sensors in the graph will be influenced. Our analysis also suggests that setting $\xi$ to an appropriate range is important. An extremely large $\xi$, which represents abnormal driving behaviors in traffic domains, can be detected easily by the user of GNNs. By analyzing PeMS and META-LA(S), speed variation within 15km/h occurs frequently, and consequently, we regard the accessible boundary of speed variation is 15 km/h, namely $\sqrt{\xi}=15$ km/h.

\begin{table*}[!t]
\centering
\small
\begin{tabular}[]{ l r r r r r r}
\toprule
\multirow{2}*{}  &\multicolumn{2}{c}{STGCN} &\multicolumn{2}{c}{DCRNN } &\multicolumn{2}{c}{Grave Wavenet }\\
%\cline{2-7}
& NMAPEI & 30\%-IV & NMAPEI & 30\%-IV & NMAPEI & 30\%-IV\\
\midrule
SFA & \textbf{15.2\%}& \textbf{17}& \textbf{16.7\%}& \textbf{22} &\textbf{15.5\%} &\textbf{21}\\
%RAN &2.1\% & 1&2.7\% & 0&2.3\% & 0\\
GWN &1.7\% & 0&2.3\% & 0&2.1\% & 0\\
DEG &4.5\% & 3&4.7\% & 3&5.7\% & 2\\
\midrule
MFGSM ($\epsilon$=2)&15.4\% & -&15.6\% & -&16.2\% & -\\
MFGSM ($\epsilon$=3)&27.3\% & -&24.4\% & -&25.8\% & -\\
\bottomrule
\end{tabular}
\caption{Effectiveness evaluation based on PeMS.}\label{table:PeMS1}
\end{table*}

%\begin{table}[ht]
%\caption{The efficiency test on the proposed weakest vertex locating strategy.}\label{table:efficiency weak}
%\begin{center}
%\begin{tabular}[]{ c c c c c }
%\hline
%\multirow{2}*{}  &\multicolumn{4}{c}{Vertex Amount}\\
%\cline{2-5}
%& 50 & 100 & 150 & 200\\
%\hline
%CT(s) & 447& 902 & 1321 & 1795\\
%Proposed(s) & 403 & 615 &  797 &1104 \\
%\hline

%\end{tabular}
%\end{center}
%\end{table}

\subsection{Effectiveness of Spatially Focused Attack}

We set parameters and hyperparameters in SFA based on the aforementioned experimental results. Finally, we conduct experiments on PeMS to show the overall the effectiveness of the proposed method. We perform 15 min-ahead traffic speed forecasting using the three mentioned GNNs and compare the proposed SFA with four attack baselines.
\begin{itemize}
     \item GWN: Generate Gaussian White Noise (GWN) with the scale being consistent with $\sqrt{\xi}$ as in SFA, and attack the same weakest sensor. This baseline is used to evaluate the effectiveness of perturbation design as in Eq.~\eqref{equ:onevertex}.
    \item DEG: We simply consider the sensor with the largest number of neighbors as the ``weakest'', and attack it using the same optimization algorithm as in SFA. This baseline is designed to examine whether the proposed locating strategy can effectively identify the weakest sensor (or a suboptimal that also works well).
    \item MFGSM: Attack all sensors with the modified Fast Gradient Sign Method (FGSM)~\cite{fgsm,moosavi2016deepfool}. We replace the ground truth in the original FGSM with the proposed inverse estimation as Eq.~\eqref{equ:ie}. The modified FGSM perturbation is computed as
    \begin{equation}\label{equ:fgsm}
        \bm{\rho}=\epsilon\ \operatorname{sign}(	\bigtriangledown \mathcal{J}(\Phi,\bm{\mathcal{G}},\tilde{\mathcal{G}}_{t})),
    \end{equation}
    where $\bigtriangledown\mathcal{J}$ computes the gradient of the cost function around the prediction of the forecasting model parameterized by $\Phi$ w.r.t the input sequence $\bm{\mathcal{G}}$, $sign$ denotes the sign function, $\tilde{\mathcal{G}}_{t}$ denotes the inverse estimation of the ground truth, and $\epsilon$ control the perturbation's scale. This baseline is chosen to compare the targeted attack in SFA with manipulating all sensors. We set the scale parameter, $\epsilon$, to 2 and 3, respectively.
\end{itemize}

% \begin{figure}[htbp]
% \begin{subfigure}{.21\textwidth}
% \centering
% \includegraphics[width=4cm]{errors2.pdf}
% \caption{Attack A.}\label{fig:attackmap}
% \end{subfigure}
% \begin{subfigure}{.23\textwidth}
% \centering
% \includegraphics[width=4cm]{newinput.pdf}
% \caption{A's input.}\label{fig:input}
% \end{subfigure}
% \begin{subfigure}{.23\textwidth}
% \centering
% \includegraphics[width=4cm]{new55.pdf}
% \caption{B's prediction}\label{fig:55}
% \end{subfigure}
% \begin{subfigure}{.23\textwidth}
% \centering
% \includegraphics[width=4cm]{new205.pdf}
% \caption{C's prediction}\label{fig:205}
% \end{subfigure}
% \caption{Results of the proposed Spatial Focusing Attack. (a) shows the NMAPEI map at timestamp 60 and only A is attacked.}
% \end{figure}

\begin{figure}[!t]
    \centering
\subfigure[Attack A]{\includegraphics[width=3.5cm]{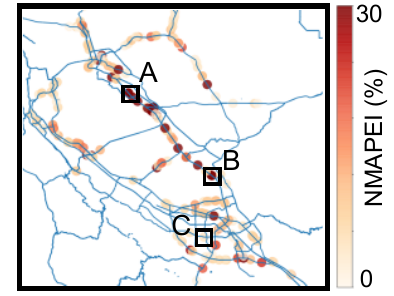}\label{fig:attackmap}}\quad
\subfigure[Input to sensor A]{\includegraphics[width=3.3cm]{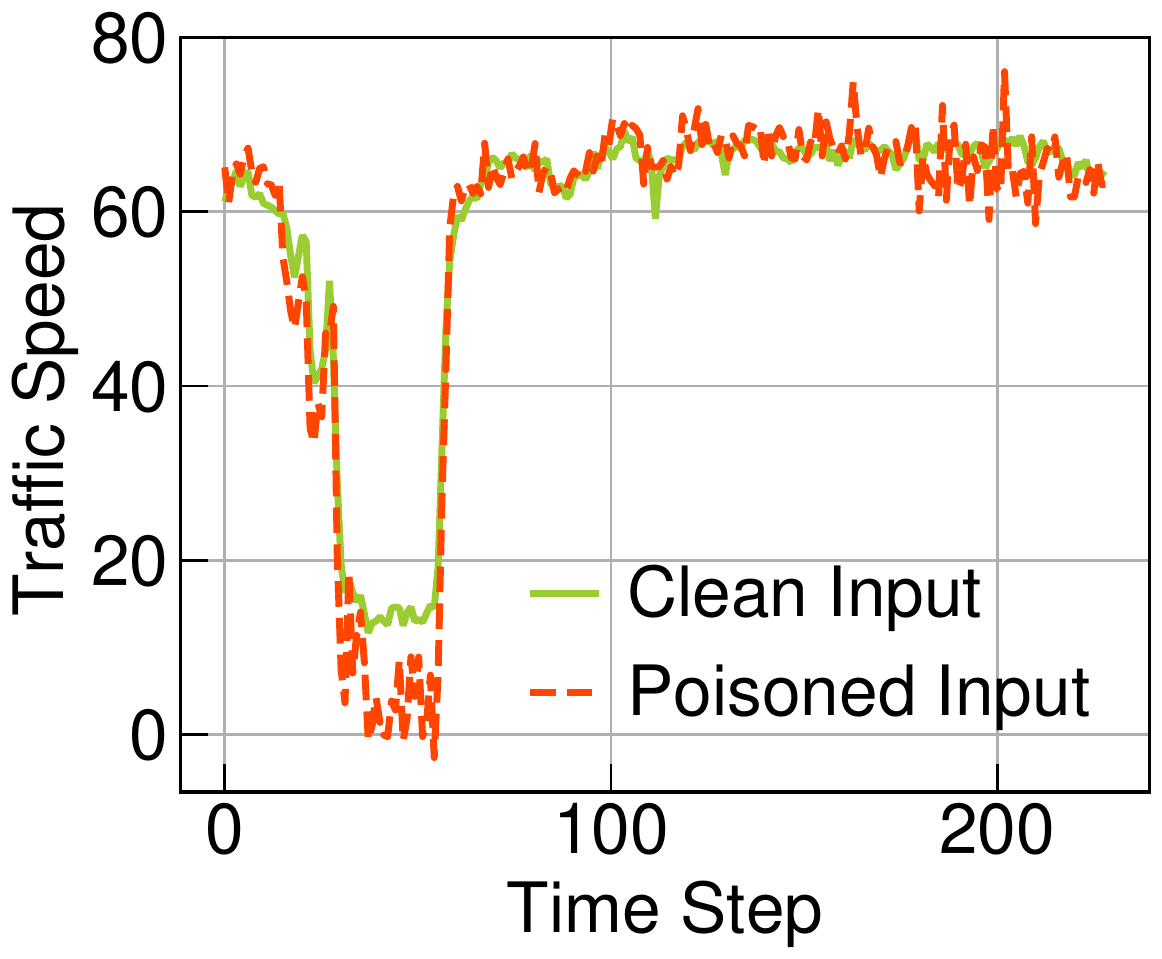}\label{fig:input}}\\
\subfigure[Prediction of sensor B]{\includegraphics[width=3.5cm]{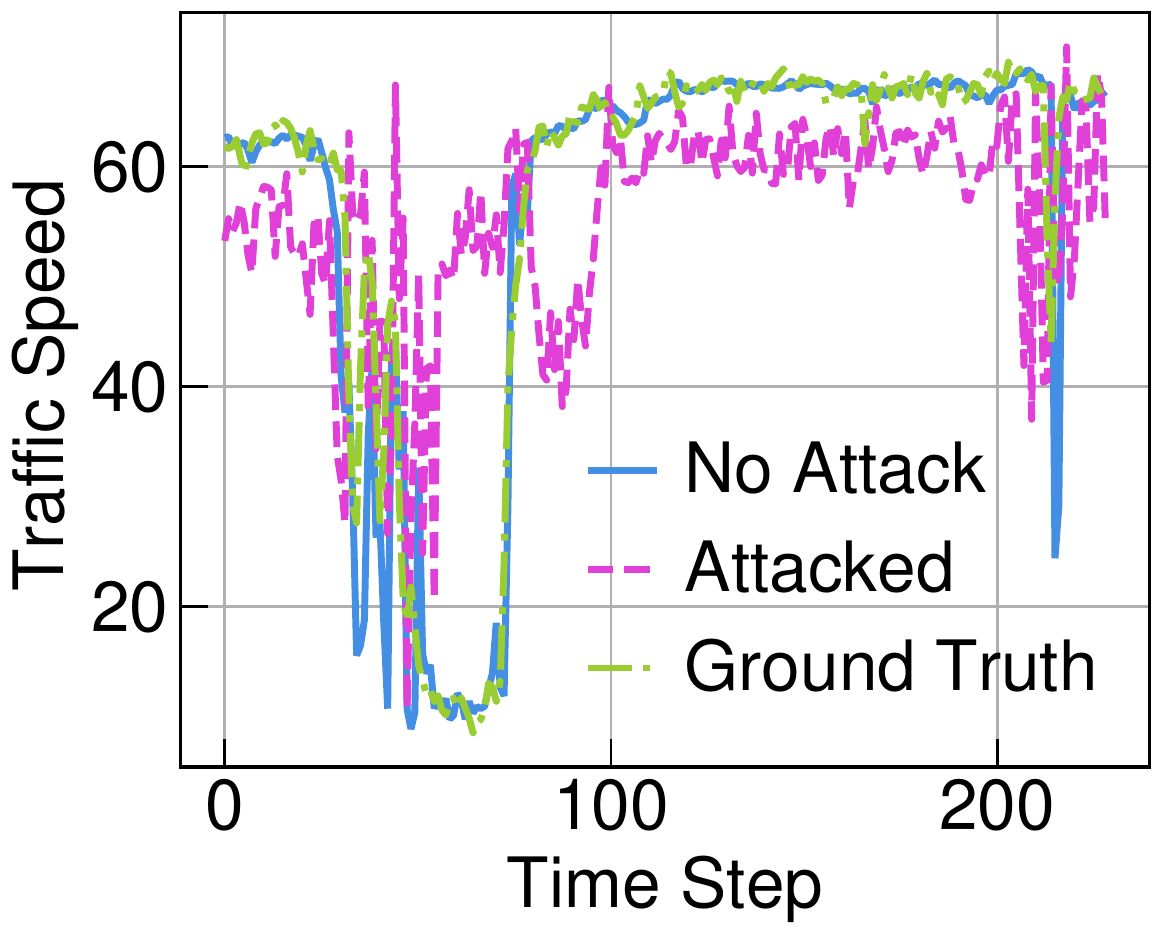}\label{fig:55}} \quad
\subfigure[Prediction of sensor C]{\includegraphics[width=3.5cm]{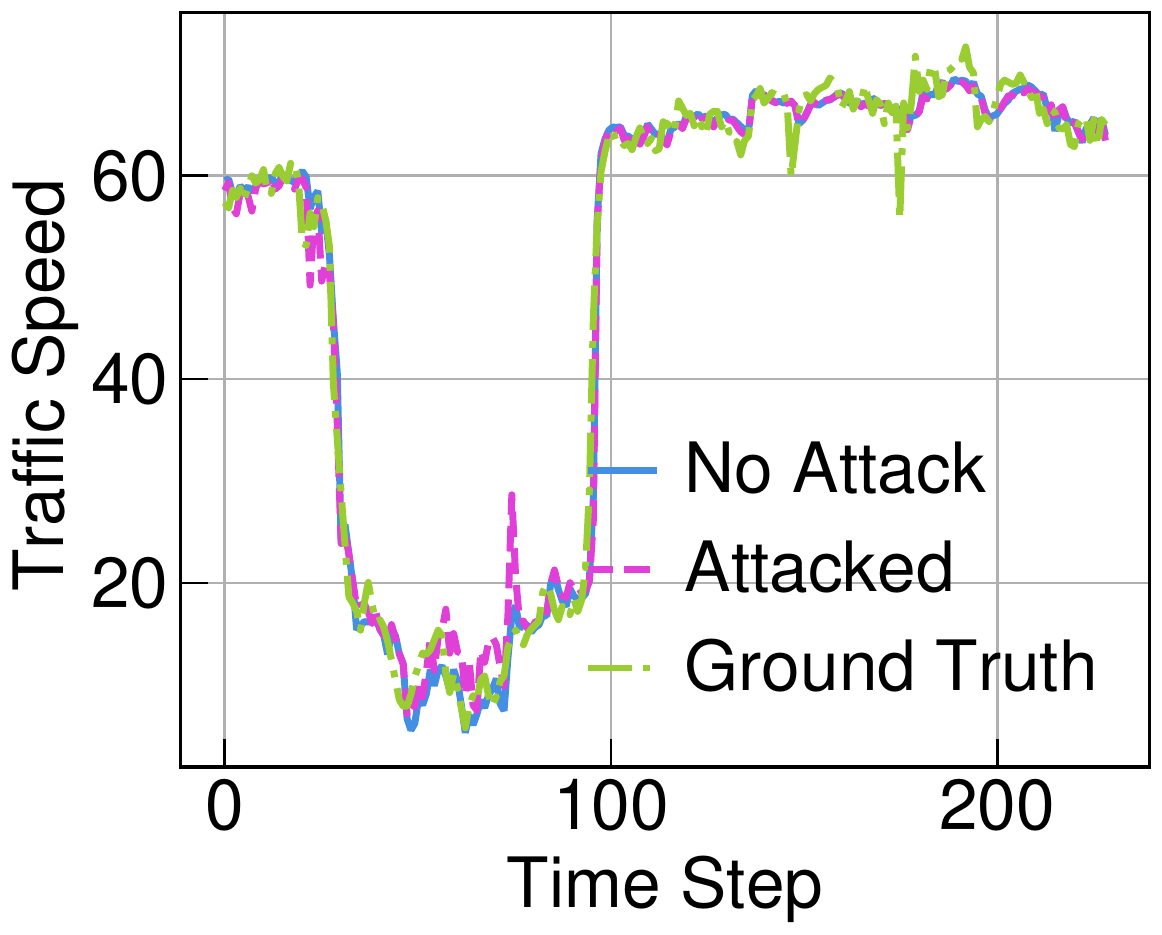}\label{fig:205}}
\caption{Results of the proposed Spatially Focused Attack. (a) The NMAPEI values for all sensors at time step 60 when A is attacked. (b) The designed perturbations to attack sensor A. (c) and (d) Attack results for sensors B/C.} \label{fig:5all}
\end{figure}

%\begin{table*}[ht]
%\caption{Effectiveness evaluation based on EI %Nino}\label{table:EINino}
%\begin{center}
%\begin{tabular}[]{ |c |c| c| c| c| c| c|}
%\hline
%\multirow{2}*{}  &\multicolumn{2}{c|}{STGCN} %&\multicolumn{2}{c|}{DCRNN } &\multicolumn{2}{c|}{Grave Wavenet %}\\
%\cline{2-7}
%& NMAPED & 30\%-IV & NMAPED & 30\%-IV & NMAPED & 30\%-IV\\
%\hline
%Proposed & \textbf{47.7\%}& \textbf{14}& \textbf{50.8\%}& %\textbf{16} &\textbf{48.1\%} &\textbf{16}\\
%RAN &5.3\% & 1&6.5\% & 2&6.0\% & 2\\
%DEG &14.5\% & 3&23.2\% & 5&18.4\% & 3\\
%\hline
%MFGSM-300&110.5\% & -&127.0\% & -&105.4\% & -\\
%MFGSM-1500&40.5\% & -&41.0\% & -&38.2\% & -\\
%\hline

%\end{tabular}
%\end{center}
%\end{table*}

In these experiments, we set $\sqrt{\xi}$ to 15 km/h for methods that attack only one sensor (i.e., SFA, GWN, and DEG). Table~\ref{table:PeMS1} shows the experiment results, which confirm the superior performance of the proposed SFA framework. We can see that SFA outperforms attacking the vertex with the most edges (DEG), showing that SFA can effectively identify a weak sensor. SFA also outperforms GWN that attacks the same weakest sensor, confirming that the proposed method can generate the optimal perturbations for manipulating only one vertex to fool the spatiotemporal forecasting model in the entire graph.

Fig.~\ref{fig:5all} shows an example of attack results on PeMS using STGCN as the forecasting model. We apply SFA to attack sensor A with optimized/designed perturbations (see Fig.~\ref{fig:input}). In Figs.~\ref{fig:55} and \ref{fig:205}, we the values/results on sensors (B) and (C), including the ground truth traffic speed, the default forecasting of STGCN without attacks, and the forecasting results when implementing the perturbations on senosr A. As can be seen, the attack on sensor A can cause substantial accuracy drop on sensor B, while sensor C is less affected by the attack. A potential reason is that A and B are connected on the highway, thus having strong dependencies, while the forecasting of sensor C might be mainly determined by its local neighbors in GNNs. Overall, for traffic forecasting in PeMS, the proposed SFA framework can cause more than 15\% accuracy drop for all three Spatiotemporal GNN models, and about 10\% sensors are severely impacted (the NMAPEI of these sensors are greater than 30\%) when setting $\sqrt{\xi}$ to 15 km/h.

Nevertheless, it should be noted that attacking all sensors will always be more effective than attacking only one vertex; however, the results in Table.~\ref{table:PeMS1} show that the one-vertex-based SFA can offer comparable performance as MFSDM that attack all sensors when setting $\epsilon= 2 \text{ km/h }\approx 0.13 \sqrt{\xi}$ (both $\sqrt{\xi}$ and $\epsilon$ control the level of perturbations). Overall, the above experiments confirms the effectiveness of SFA. The perturbation can be diffused into the entire graph and even predictions on sensors that are far from the attacked one can be severely influenced.

%the proposed method's effectiveness is similar with attacking all sensors by MFSDM with the perturbation scale, $\epsilon$, approximately equal to $10\%$ of the SFA's perturbation scale, $\sqrt{\xi}$, which is concluded by comparing the proposed method with MFGSM($\epsilon$=2) and MFGSM(($\epsilon$=2)) in table~\ref{table:PeMS1}.

%It should be noted that MFGSM attacks 200 sensors, while the proposed SFA only attacks one.
%\subsection{Trade-off between Effectiveness and Efficiency}
%{\color{red}delet it?}

%In subsection~\ref{subsection:weakest}, we design a genetic algorithm-based method to locate the weakest vertex in a graph rather than applying complete traversal algorithm. In this subsection, we test how the parameter $s$, which determine how many vertices are selected as candidates, impacts the proposed one vertex attacking method's effectiveness and efficiency. This test is based on PeMS, the SNR is set as 10, and STGCN~\cite{yu2017spatio} works as the target forecasting model. We set $s$ differently and then compare the computation time and NMAPED, which is shown as Table~\ref{table:tradeoff}. It is concluded that the effectiveness will not keep increasing with the increase of $s$, and the efficiency will not keep becoming worse.

%\begin{table*}[ht]
%\caption{Trade-off Study}\label{table:tradeoff}
%\begin{center}
%\begin{tabular}[]{ |c |c| c| c| c| c|}
%\hline
%& 5& 10& 15& 20&25\\
%\hline
%Time(s) & 3.3& 6.4& 8.7& 8.9& 9.0\\
%NMAPED &15.2\% & 19.3\% & 19.6\%& 19.6\%&19.6\%\\
%\hline

%\end{tabular}
%\end{center}
%\end{table*}

\section{Conclusion}

In this paper, we propose Spatially Focused Attack (SFA) to break GNNs-based spatiotemporal forecasting models by poisoning only one vertex/sensor. SFA consists of three key components---using inverse estimation to effectively design a universal perturbation, identifying the most vulnerable sensor based on ``weakness'', and redesign perturbations on the selected sensor. Different from other attack studies, the routine of SFA does not require future information in computing the optimal perturbations. Our experiments on two real-world traffic datasets have demonstrated the effectiveness of the single sensor-based attacks. One direction for future research is to seek solutions to reformulate the white-box attack to generate black-box perturbations. Given that SFA has shown that attacking a single sensor can cause network disruption, how to defend the real-world forecasting systems and to make them more robust is an urgent research question for agencies and practitioners.

%Future works for the proposed study can be summarized as follows. First, we utilized a universal adversarial attack method to measure the ``weakness" of vertices. We do not include temporal patterns in our measurement. Consequently, involving temporal patterns in the evaluation is a possible modification. Second, we design a genetic algorithm-based method to find the ``weakest" vertex in a graph to attack. This might not be the optimal solution. Third, studies on the scalability of one vertex attack is valuable.

%Besides, as spatiotemporal applications require reliable algorithms, how to defend these adversarial attacks, and how to build more robust spatiotemporal GNNs-based models are still valuable.

%\bibliographystyle{named}
\bibliography{aaai22}

\end{document}